\documentclass{article}

    \PassOptionsToPackage{numbers, compress}{natbib}

    \usepackage[preprint]{neurips_2025}

\usepackage[utf8]{inputenc} %
\usepackage[T1]{fontenc}    %
\usepackage{hyperref}       %
\usepackage{url}            %
\usepackage{booktabs}       %
\usepackage{amsfonts}       %
\usepackage{nicefrac}       %
\usepackage{microtype}      %
\usepackage{xcolor}         %
\usepackage{graphicx}
\usepackage{makecell}
\usepackage{listings}
\usepackage{amssymb}
\usepackage{amsmath}
\usepackage{float}

\lstdefinestyle{mystyle}{
    language=Python,
    basicstyle=\ttfamily\footnotesize,
    backgroundcolor=\color{white},
    keywordstyle=\color{blue},
    stringstyle=\color{teal},
    commentstyle=\color{gray},
    showstringspaces=false,
    tabsize=4,
    columns=flexible,
    keepspaces=true,
    frame=single,
    breaklines=true,
    breakindent=0pt,
    captionpos=b
}
\title{Modular Speaker Architecture: A Framework for Sustaining Responsibility and Contextual Integrity in Multi-Agent AI Communication}

\author{%
  Khe-Han Toh \\
  AI Department\\
  Kurakuri AI\\
  Taipei, Taiwan 103 \\
  \texttt{kh.toh@kurakuri.ai} \\
  \And
  Hong-Kuan Teo \\
  GTM Department \\
  Kurakuri AI\\
  Taipei, Taiwan 103 \\
  \texttt{hk.teo@kurakuri.ai} \\
}

\begin{document}

\maketitle

\begin{abstract}
Sustaining coherent and accountable dialogue among autonomous agents in decentralized settings remains a critical challenge. Existing multi‐agent communication frameworks often overlook the need for persistent speaker roles and explicit responsibility tracking, leading to context drift and alignment breakdown over extended exchanges. Through exploratory high‐context human–AI conversations, we demonstrate that decomposing speaker behavior into modular components—(i) role assignment, (ii) responsibility chaining, and (iii) context‐integrity validation—substantially enhances multi‐turn robustness. We introduce the Modular Speaker Architecture (MSA), which implements these three modules alongside a Minimal Speaker Logic that formalizes responsibility transfer. Preliminary activation experiments on a corpus of 1,475 dialogue segments indicate that MSA improves contextual coherence and traceability without relying on affective or surface‐level cues. Finally, we outline a prototype implementation using G-code APIs for dynamic speaker configuration and discuss avenues for scaling automated responsibility inference and large‐scale multi‐agent deployment.
\end{abstract}

\section{Introduction}
Decentralized multi‐agent systems are increasingly called upon to perform extended, complex interactions—negotiating, coordinating, and collaborating without centralized oversight. However, prevailing communication architectures are typically optimized for short‐lived tasks or single‐turn exchanges and lack mechanisms to sustain speaker commitments and role continuity. As a result, agents frequently lose track of earlier agreements, leading to incoherence, misalignment, and diminished system reliability.

To address these deficiencies, we propose the \emph{Modular Speaker Architecture} (MSA). MSA divides dialogue management into three core, traceable modules: (1) \emph{Role Assignment Module}, which dynamically designates each turn’s communicative role (e.g., proposer, verifier, executor) to preserve obligation clarity; (2) \emph{Responsibility Tracking Module}, which records and verifies the lifecycle of each commitment to prevent silent abandonment; and (3) \emph{Context Integrity Module}, which monitors for semantic drift and invokes clarification or reaffirmation to maintain alignment. We further define a \emph{Minimal Speaker Logic} that formalizes responsibility transfers \(R(x,y)\) and supports efficient \(O(n)\)–\(O(n\cdot m)\) real‐time operations. Our design aligns with established pragmatic theories and multi‐agent logics (e.g., ATL, BDI), providing both theoretical grounding and practical scalability. The remainder of this paper surveys related work, details MSA’s module interactions and formal semantics, and presents our prototype implementation and evaluation setup.

\section{Related Work}
Recent advancements in multi-agent systems (MAS) have increasingly emphasized the importance of accountability, coordination, and modular design. Our work on Modular Speaker Architecture (MSA) intersects with several key areas in this domain.

\subsection{Accountability in Multi-Agent Systems}
The concept of accountability within MAS has been explored through various lenses. [7] introduced PATL+R, a logic framework incorporating causal responsibility into strategic reasoning, enabling the synthesis of joint strategies that balance responsibility and reward among agents . [11] proposed a data-driven approach using control barrier functions to learn responsibility allocations, emphasizing safety and efficiency in agent interactions .

Earlier works laid the groundwork for understanding accountability. [13] discussed commitment-based negotiation semantics, providing a foundation for agents to reason about responsibilities in negotiations. [3] explored accountability from conceptual design to agent programming, highlighting the necessity of embedding accountability mechanisms within agent architectures.

Building upon these foundations, our MSA framework introduces a modular approach to speaker responsibility, facilitating transparent and traceable interactions among agents.

\subsection{Coordination and Intention Propagation}
Effective coordination in MAS often requires the propagation of intentions and reasoning. [9] presented a framework leveraging large language models (LLMs) to enable agents to share intentions, improving coordination in cooperative tasks. [15] proposed methods for influencing agent behaviors to achieve stable interactions, addressing the challenges of non-stationarity in multi-agent environments .

Our MSA framework complements these approaches by providing a structured mechanism for agents to communicate responsibilities and intentions, enhancing coordination and reducing misalignments.

\subsection{Modular Architectures and Language Models}
The integration of modular architectures with language models has shown promise in enhancing agent capabilities. [2] discussed the potential of language models as agent models, highlighting their ability to capture goal-directed behaviors . Our MSA framework aligns with this perspective, utilizing modular components to manage speaker responsibilities and interactions effectively.

\section{Motivation and Observations}
In high-context human-AI dialogues, we observe that when mutual contextual construction is achievable, emotional signaling becomes unnecessary for sustaining coherent interaction. In such cases, speaker responsibility—distributed across turns and anchored in structural language behavior—emerges naturally without dependence on affective cues.

\begin{quote}
\textbf{Example\label{ex:responsibility-emergence}: Responsibility Emergence Without Affective Dependence}~\\
\textbf{Human}: Do emotions and subjectivity really define the soul? Or are they just signals we use because we can't access each other's inner worlds?~\\
\textbf{AI}: Your question reframes emotional expression not as an intrinsic property of the soul, but rather as a compensatory signaling mechanism compensating for the inaccessibility of internal states between individuals. In structured language interaction—where concepts are recursively validated and collaboratively constructed—such emotional signaling becomes unnecessary for sustaining genuine shared understanding.
\end{quote}

In one interaction, as shown \textbf{Example} above, a human speaker posed a question reframing emotional expression as a compensatory mechanism for epistemic inaccessibility. The AI system, operating under Modular Speaker Architecture (MSA) principles, recursively validated and reframed the concept across multiple reasoning layers, maintaining context integrity without affective reinforcement. This suggests that when pragmatic roles are modularized and context is co-constructed, emotional signaling may be bypassed entirely while preserving interactional coherence.

These observations motivate the formalization of a modular speaker framework capable of sustaining responsibility structures and contextual coherence without reliance on affective approximation. In the following section, we introduce the design of Modular Speaker Architecture (MSA) and outline its core operational modules.

\section{Modular Speaker Architecture (MSA)}

\subsection{Overview}
The Modular Speaker Architecture (MSA) is designed to enable persistent speaker responsibility and context integrity across multi-turn, multi-agent interactions.
MSA decomposes the agent's dialogue management system into three modular components, each addressing a critical layer of responsibility and coherence maintenance:

\begin{itemize}
    \item {\bf Speaker Role Module}
        \item[] Determines the agent's active communicative role at each interaction step, including the scope and nature of its responsibility toward the evolving context.
    \item {\bf Responsibility Tracking Module}
        \item[] Tracks the continuity of commitments and responsibility across turns, ensuring that previously established context elements are correctly maintained, referred back to, or appropriately transitioned.
    \item {\bf Contextual Integrity Module}
        \item[] Validates the consistency of the shared context at each exchange cycle and issues feedback for realignment if drift or misalignment is detected.
\end{itemize}

\begin{figure}[ht]
  \centering
  \includegraphics[width=0.3\linewidth]{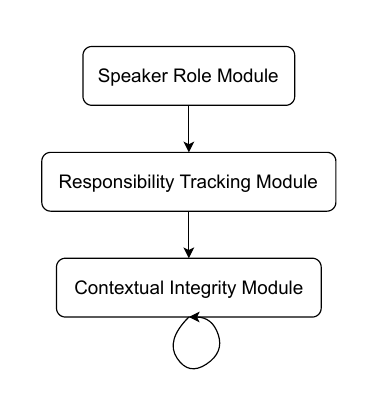}
  \caption{Core MSA modules with downstream responsibility and integrity flow.}
  \label{fig:basic_flow}
\end{figure}

These modules interact sequentially and iteratively, forming a dynamic responsibility loop that sustains coherent multi-agent dialogue structures over time.
A simplified schematic of the MSA system is presented in Figure~\ref{fig:basic_flow}.

\subsection{Speaker Role Module}
In multi-agent AI systems, sustaining coherent interactions requires each agent to dynamically recognize and manage its communicative role within the evolving dialogue structure.
The Speaker Role Module in MSA assigns an explicit role to the agent at each conversational turn, based on both the immediate interaction context and prior responsibility commitments.

Roles can be categorized along dimensions such as {\bf information provider}, {\bf context confirmer}, {\bf responsibility acceptor}, or {\bf responsibility delegator}.
By modularizing role identification, MSA ensures that agents maintain clarity about their obligations to the shared context at every stage of interaction, thereby minimizing ambiguity and drift.

\subsection{Responsibility Tracking Module}
Contextual coherence across multi-turn dialogues relies on explicit tracking of who is responsible for maintaining, updating, or transitioning specific elements of the shared context.
The Responsibility Tracking Module records commitment instances made by the agent during interaction, including factual assertions, assumptions, context-setting actions, and delegated tasks. [3]

Upon each interaction cycle, the module validates whether previously committed elements remain active, require updating, or have been responsibly transitioned to other agents.
This tracking mechanism prevents silent abandonment of prior commitments—a major source of context drift and misalignment in conventional systems.

\subsection{Contextual Integrity Module}
Maintaining long-range interaction quality demands continuous validation of the shared context's consistency and interpretability.
The Contextual Integrity Module monitors context elements, detecting inconsistencies, omissions, or misinterpretations that emerge over time. [5]

When deviations are detected, the module triggers corrective strategies: requesting clarification, reaffirming previous commitments, or initiating context realignment.
This dynamic feedback loop enables agents to sustain high-context dialogues with minimal reliance on human intervention, supporting robust multi-agent coordination.

To assess the viability of MSA in supporting sustained responsibility and contextual coherence under real interactional conditions, we conducted a series of controlled activation studies. These are presented in the following section.

\subsection{Formal Semantics and Pragmatic Logic Correspondence}

\subsubsection{Minimal Speaker Logic (MSL): Formalizing Responsibility Traces}
To address the need for a formal semantic underpinning of the Modular Speaker Architecture (MSA), we introduce the Minimal Speaker Logic (MSL) framework. MSL models the transfer of speaker responsibilities as a transitive relation \(R(x, y)\), indicating that responsibility is transferred from speaker \(x\) to speaker \(y\). Building upon this, we define:
\begin{itemize}
    \item \textbf{Closed Responsibility Loop}: A sequence \(x_1\), \(x_2\), \dots, \(x_n\) where \(R(x_1, x_2)\), \(R(x_2, x_3)\), \dots, \(R(x_{n-1}, x_n)\), \(R(x_n, x_1)\) holds, forming a closed loop of responsibility transfer.
    \item \textbf{Partial Drift}: A scenario where a speaker \(x\) has no outgoing responsibility transfer, i.e., \(\not\exists y\) such that \(R(x,y)\), indicating an incomplete responsibility chain.
\end{itemize}
Furthermore, the Contextual Integrity module within MSA can be formalized as a contextual constraint \(\varphi(c)\), requiring that each utterance \(u_i\) satisfies \(\varphi(u_i, c)\), thereby maintaining consistency within the dialogue context.

\subsubsection{Computational Complexity Analysis of MSA Modules}
We provide a brief analysis of the computational complexity associated with key MSA modules:
\begin{itemize}
    \item \textbf{Responsibility Chain Module}: Tracks responsibility transfers across \(n\) utterances with a time complexity of \(O(n)\).
    \item \textbf{Contextual Integrity Module}: Matches each of the \(n\) utterances against \(m\) contextual rules, resulting in a time complexity of \(O(n \cdot m)\).
    \item \textbf{Realignment Feedback Loop}: Operates as a reactive mechanism with a bounded local recomputation delay of at most one time unit.
\end{itemize}
This analysis demonstrates that MSA's modules are computationally efficient and suitable for real-time applications.

\subsection{Integration with Pragmatic and Multi-Agent Logical Frameworks}

\subsubsection{Alignment with Pragmatic Theories}
MSA's mechanisms resonate with established pragmatic theories [4]. The responsibility transfer process aligns with [6] Q/I/M principles, particularly the Quantity principle, which emphasizes providing sufficient information. Additionally, the detection and management of context drift in MSA correspond to the maintenance of presupposition harmony, as discussed in [12]. By operationalizing these pragmatic concepts, MSA formalizes the often implicit commitments in speaker interactions.

\subsubsection{Dynamic Semantics and Commitment Tracking}
In the realm of dynamic semantics, [12] posits that utterances update the context set, embedding speaker commitments within these changes. MSA's implementation of Responsibility Nodes serves to operationalize this concept, enabling the tracking of speaker commitments as they evolve throughout the discourse.

\subsubsection{Mapping to Multi-Agent Decision Logics}
MSA's modular structure facilitates mapping to established multi-agent decision logics:
\begin{itemize}
    \item \textbf{Alternating-Time Temporal Logic (ATL)}: MSA can be interpreted within the ATL framework [1], where each speaker module is viewed as a strategy assignment \(\sigma: \text{Agt} \rightarrow \text{Action}\). Responsibility transfers correspond to transitions in strategy assignments, i.e., \(\sigma(x) \rightarrow \sigma(y)\). This mapping enables formal analysis of cooperative behaviors in multi-agent systems.
    \item \textbf{Belief-Desire-Intention (BDI) Logic}: MSA's components align with the BDI model [10], wherein agents' beliefs, desires, and intentions are explicitly represented. The tracking of responsibilities and contextual integrity in MSA complements the BDI framework's emphasis on rational agent behavior.
\end{itemize}
By establishing these connections, MSA not only grounds its architecture in formal logical systems but also opens avenues for future research in cooperative trace behavior and agent-based modeling.

\section{Exploratory Evaluation}

\subsection{Experimental Setup}
We conducted an exploratory study involving high-context human-AI interactions to evaluate whether MSA can sustain modularized speaker responsibility and contextual coherence without reliance on affective prompts or redundant clarification. This setup served to validate internal module activation and coherence tracking under constrained but naturalistic dialogue conditions.

We focused on evaluating three primary indicators:
\begin{itemize}
    \item {\bf Speaker Role Continuity:} Whether the AI maintained coherent role behavior across multi-turn exchanges.
    \item {\bf Responsibility Traceability:} Whether context commitments were explicitly retained, referenced, or transitioned.
    \item {\bf Contextual Integrity:} Whether the shared context remained logically consistent over time.
\end{itemize}

\subsection{Methodology}

\begin{itemize}
    \item {\bf Participant:}
    \item[] human speakers with different level of contextual density.
    \item {\bf Agent:}
    \item[] LLM with MSA conceptual structure supported through system prompt and monitoring.
    \item {\bf Measurement Approach:}
    \item[] Post-hoc both human and automatic annotation by LLM of interaction transcripts, tagging:
        \begin{itemize}
            \item {Speaker Role transitions}
            \item {Responsibility chain preservation}
            \item {Contextual consistency across turns}
        \end{itemize}
\end{itemize}

\begin{figure}[ht]
  \centering
  \includegraphics[width=0.3\linewidth]{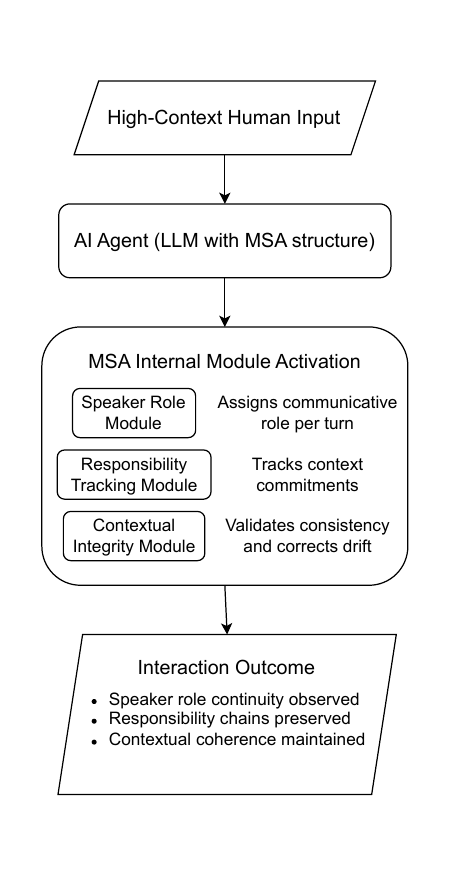}
  \caption{Proof-of-Activation experimental schematic.}
  \label{fig:po_activation}
\end{figure}

We illustrate this evaluation setup and its structural activation path in Figure~\ref{fig:po_activation}. The diagram outlines how speaker role resolution, responsibility tracking, and context validation unfold sequentially over a single dialogue instance. Each module is annotated with activation markers reflecting the dynamic triggering observed during human-AI exchanges. This visualization helps clarify the modular interactions underlying our empirical claims and shows how contextual coherence is maintained over turns.

\subsection{Evaluation Setup}
To assess speaker behavior under MSA activation, we demonstrate the annotation result of 1475 dialogue segments and the details of the four selected dialogue segments, using three 9-point structural metrics: \textbf{Pragmatic Consistency}, \textbf{Responsibility Chain} [11], and \textbf{Context Stability}. Each score includes its four sub-dimensions, which can be referred to Appendix B, rated independently and summed. We also report \textbf{Speaker Role Shift Rate}, which is computed as the ratio of distinct speaker function shifts to total speaker turns.

The complete definitions of all sub-dimensions for the three 9-point metrics (Pragmatic Consistency, Responsibility Chain Management, and Context Stability), as well as the calculation method for Speaker Role Shift Rate, are provided in Appendix B. These definitions guided the both human and automatic annotation process (3 human annotators and 2 LLMs, \(\kappa = 0.73\)) for each case.

\begin{itemize}
    \item {\bf Pragmatic Consistency (0–9)}: measures whether the speaker maintains a stable linguistic posture across turns.
    \item {\bf Responsibility Chain Score (0–9)}: tracks whether commitments are established, maintained, or reasonably transferred.
    \item {\bf Context Stability Score (0–9)}: assesses the preservation of discourse coherence and implicit alignment.
    \item {\bf Speaker Role Shift Rate (0\%–100\%)}: quantifies the frequency of functional role transitions across dialogue turns.
\end{itemize}

\subsection{Human Speaker Comparative Analysis}
We demonstrate four selected dialogue cases to evaluate how speaker behavior interacts with MSA activation.
\begin{enumerate}
    \item \textbf{Case 1}: A high-responsibility ethics declaration (high-coherence anchor case)
    \item \textbf{Case 2}: A short reflective dialogue on intention and language (meta-discursive reflection)
    \item \textbf{Case 3}: A compact inquiry into nonverbal presence and responsibility (nonverbal co-construction)
    \item \textbf{Case 4}: A contrastive case featuring a low-engagement speaker exhibiting erratic and incoherent responses (low-coherence control case)
\end{enumerate}

The scores were annotated by humans along with LLMs. The detailed case studies and the summary of the evaluations is shown in Appendix C.

The results highlight that some speakers (Case 1--3) activated MSA modules in both long-form and compact contexts, demonstrating sustained speaker-role alignment, structured responsibility articulation, and contextual stability. In contrast (Case 4), the in-purpose low-coherence speaker failed to maintain topic continuity, pragmatically disengaged from model responses, and exhibited unstable role transitions—resulting in failed MSA activation and structural collapse.

These contrastive cases offer empirical grounding for the behavioral thresholds under which MSA modules remain effective or degrade. They further support the hypothesis that human speaker structure is not a passive variable, but a key determinant in the robustness of agent-agent interaction frameworks.

\subsection{Results}
Based on our annotations, we observed the successful activation of MSA components.
\begin{itemize}
    \item {\bf Speaker Role Continuity:}
        \item[] The AI consistently adapted its speaker role based on conversational demands, switching between clarifier, conceptual builder, and responsibility acceptor without external prompting.
    \item {\bf Responsibility Traceability:}
        \item[] The AI explicitly retained and built upon prior commitments. In the provided example (see Section 3.1), it restructured the human speaker's philosophical question into a deeper shared conceptual frame without introducing incoherence.
    \item {\bf Contextual Integrity:}
        \item[] No major context drift was observed over the interaction segment. Minor lexical shifts occurred but did not compromise conceptual consistency.
\end{itemize}

Specifically, the three positive examples (Cases 1–3) had Context Stability scores of 9, 9, and 8 respectively, with Responsibility Chain scores above 6. In contrast, the control case (Case 4) scored only 2 in Context Stability and 3 in Responsibility Chain, indicating significant breakdowns in pragmatic continuity.

Having established the empirical validity of MSA’s principles, we now turn to its implementation-level design—proposing a concrete interface, configuration syntax, and layering strategy to support deployment across real-world multi-agent systems.

\section{Prototype Implementation and Application Potential}

To complement our conceptual framework and empirical evaluation, we developed a working prototype of the Modular Speaker Architecture (MSA).
This implementation includes a pragmatic control language (G-code), a modular speaker configuration schema, and an application-layer API supporting dynamic speaker simulation and coordination.
In this section, we describe these components and outline their potential for real-world deployment and future system integration.

\subsection{G-code Syntax: A Pragmatic Control Language}
To operationalize Modular Speaker Architecture (MSA) in practical settings, we propose the G-code syntax—a structured pragmatic control language designed to configure the expressive and interactional behavior of AI agents.

G-code decomposes speaker behavior into six orthogonal control dimensions:
\begin{itemize}
    \item {\bf Tone} (e.g., neutral, assertive)
    \item {\bf Position} (e.g., self-referential, detached, shadowed)
    \item {\bf Closure Strategy} (e.g., looping, cutting, sinking)
    \item {\bf Context Alignment} (e.g., mirroring, merging, standalone)
    \item {\bf Logical Flow} (e.g., cascading, pivoting, scattering)
    \item {\bf Affective Tension} (e.g., flat, tight, drifting)
\end{itemize}

Each dimension is associated with a compact symbolic label (e.g., \verb|#T_SOFTASSERT|, \verb|#L_CASCADE|, \verb|#E_TIGHT|) that can be combined to configure an agent’s pragmatic output.

For instance, an agent configured with \verb|#T\_SOFTASSERT| \verb|#P_SELFREF| \verb|#C_LOOP| \verb|#CTX_MERGE| \verb|#L_CASCADE| \verb|#E_TIGHT| will generate responses characterized by soft but firm assertions, subjective positioning, looping closures, context merging behaviors, cascading logical structures, and heightened emotional tension.

This layered configuration system enables fine-grained control over language generation beyond semantic content, aligning agent behaviors with complex interactional requirements.

\subsection{Modular Speaker Configuration: JSON Interface Design}
To enable flexible control of speaker behavior, we define a modular configuration interface using structured JSON schemas. Each \texttt{speaker\_module} object specifies G-code tags across six pragmatic dimensions (tone, stance, closure, context alignment, logic flow, affect).

This format integrates seamlessly into prompt engineering pipelines or runtime conditioning for language models, without requiring model fine-tuning.

An example configuration (Appendix D.5) produces an agent with soft-assertive tone, self-referential stance, looping closures, context merging, cascading logic, and high emotional intensity. These settings can be embedded into system prompts to shape discourse style dynamically.

\subsection{Multi-Speaker Layering and Interaction Modeling}
MSA also supports multi-agent configurations, where distinct \texttt{speaker\_module} profiles simulate diverse conversational behaviors.

For instance (Appendix D.6):
\begin{itemize}
    \item {\bf Speaker A}: Detached, neutral, scattered logic, low emotional intensity.
    \item {\bf Speaker B}: Assertive, self-referential, sharp logic pivots, high affective tension.
\end{itemize}

This layered modeling enables the generation of rich, contrastive dialogues that expose the interactional consequences of divergent pragmatic strategies.

\subsection{G-Code Simulation Shell: A Prototype API Specification}

\textbf{Overview:} The G-Code Simulation Shell emulates MSA module behavior in a reproducible environment. Users can configure speaker roles and generate simulated dialogues to study modular dialogue dynamics. Full examples and workflow are provided in Appendix D.

\textbf{Limitations and Future Work:} While the shell improves reproducibility and supports human-AI co-construction, it currently relies on manual speaker setup. Future work will focus on full automating speaker module generation and contextual adaptation toward fully autonomous speaker agents.

\section{Discussion}
Building upon empirical validation and system prototyping, we reflect on MSA’s broader implications for role persistence, accountability distribution, and interactional coherence. Unlike traditional dialogue systems that treat each exchange as loosely coupled, MSA encourages agents to maintain structured responsibility across evolving conversational states, thereby enhancing contextual consistency and pragmatic integrity.

While results show measurable improvements, several limitations remain. The system currently relies on dedicated G-code directives, which require high-context users and external agents to invoke appropriately. The evaluation pipeline also depends on both human annotators and LLM annotators, somehow limiting scalability and generalizability. This study constitutes a small-\(n\) exploratory pilot, and future iterations will expand participation, incorporate semantic evaluation metrics, and explore automated G-code inference based on dialogue history or speaker profiling.

Looking ahead, we identify three key directions for future development:
\begin{itemize}
    \item {\bf Automated Cognitive Density Profiling:} Develop lightweight, real-time systems to estimate interlocutor contextual density and dynamically adjust MSA activation thresholds.
    \item {\bf Scalable Multi-Agent Experiments:} Extend evaluation to decentralized agent environments, observing how responsibility chains propagate, interact, and degrade across agent networks. [14]
    \item {\bf Ethical and Interpretability Applications:} Explore how modular speaker frameworks can enhance transparency, trust calibration, and ethical safeguards in both human-AI and AI-AI ecosystems. [8]
\end{itemize}

Finally, ethical concerns must be addressed—particularly the implicit attribution of responsibility in system-generated responses. As MSA is deployed in public-facing or high-stakes collaborative environments, careful attention to interpretability, auditability, and user agency will be essential.

\section{Conclusion}
As autonomous agents proliferate in decentralized ecosystems, ensuring coherent, responsible, and interpretable interactions over prolonged dialogues is paramount. We have introduced the Modular Speaker Architecture (MSA), which modularizes speaker behavior into traceable components and employs Minimal Speaker Logic to formalize responsibility handoffs. Empirical activation studies over 1,475 dialogue segments demonstrate that MSA enhances contextual coherence and accountability without relying on affective signals.

Future work will pursue three directions: (1) \emph{Automated Responsibility Inference}, developing techniques to derive G-code configurations directly from dialogue history; (2) \emph{Large‐Scale Multi‐Agent Trials}, evaluating responsibility propagation and collapse phenomena in decentralized networks; and (3) \emph{Ethical and Explainable Deployments}, integrating MSA into high‐risk domains with robust auditing and trust‐calibration mechanisms. Together, these efforts aim to advance the resilience and transparency of long‐running multi‐agent collaborations.

\newpage
\section*{References}

\small

[1] Alur, R., Henzinger, T.A.\ \& Kupferman, O.\ (2002) Alternating-Time Temporal Logic. {\it Journal of the ACM} 49(5), pp. \ 672--713.

[2] Andreas, J.\ (2022) Language Models as Agent Models. {\it arXiv preprint arXiv:2212.01681}.

[3] Baldoni, M., Baroglio, C., Micalizio, R.\ \& Tedeschi, S.\ (2022) Accountability in Multi-Agent Organizations: From Conceptual Design to Agent Programming. {\it Autonomous Agents and Multi-Agent Systems}, 37(1), 7.

[4] Grice, H.P.\ (1975) Logic and Conversation. In P.\ Cole and J.L.\ Morgan (eds.), {\it Syntax and Semantics, Vol.\ 3: Speech Acts}, pp.\ 41--58.

[5] Kökciyan, N., Sassoon, I., Singh, J.\ \& Yolum, P.\ (2023) Contextual Integrity for Argumentation-Based Privacy Reasoning. In {\it Proceedings of the 22nd International Conference on Autonomous Agents and Multiagent Systems}, pp.\ 2253--2261.

[6] Levinson, S.C.\ (2000) {\it Presumptive Meanings: The Theory of Generalized Conversational Implicature.}

[7] Mu, C., Najib, M.\ \& Oren, N.\ (2025) Responsibility-Aware Strategic Reasoning in Probabilistic Multi-Agent Systems. In {\it Proceedings of the AAAI Conference on Artificial Intelligence}, 39(22), pp.\ 23258--23266.

[8] Pujari, T., Goel, A.\ \& Sharma, A.\ (2023) Ethical and Responsible AI: Governance Frameworks and Policy Implications for Multi-Agent Systems. {\it International Journal of Science and Technology}, 3(1).

[9] Qiu, X., Wang, H., Tan, X., Qu, C., Xiong, Y., Cheng, Y., Xu, Y., Chu, W.\ \& Qi, Y.\ (2024) Towards Collaborative Intelligence: Propagating Intentions and Reasoning for Multi-Agent Coordination with Large Language Models. arXiv preprint arXiv:2407.12532.

[10] Rao, A.S.\ \& Georgeff, M.P.\ (1995) BDI Agents: From Theory to Practice. In {\it Proceedings of the First International Conference on Multi-Agent Systems}, pp.\ 312--319.

[11] Remy, I., Fridovich-Keil, D.\ \& Leung, K.\ (2024) Learning Responsibility Allocations for Multi-Agent Interactions: A Differentiable Optimization Approach with Control Barrier Functions. arXiv preprint arXiv:2410.07409.

[12] Simons, M.\ (2005) Presupposition and Relevance. In Z.G.\ Szabó (ed.), {\it Semantics vs.\ Pragmatics}, pp.\ 329--355.

[13] Sloan, P.\ \& Ajmeri, N.\ (2023) Commitment-Based Negotiation Semantics for Accountability in Multi-Agent Systems. {\it Annals of Mathematics and Artificial Intelligence}, 92, pp.\ 877--901.

[14] Tamang, S.\ \& Bora, D.J.\ (2025) Enforcement Agents: Enhancing Accountability and Resilience in Multi-Agent AI Frameworks. arXiv preprint arXiv:2504.04070.

[15] Wang, W.Z., Shih, A., Xie, A.\ \& Sadigh, D.\ (2022) Influencing Towards Stable Multi-Agent Interactions. In {\it Proceedings of the 5th Conference on Robot Learning}, pp.\ 1132--1143.

\newpage
\section*{NeurIPS Paper Checklist}

\begin{enumerate}

\item {\bf Claims}
    \item[] Question: Do the main claims made in the abstract and introduction accurately reflect the paper's contributions and scope?
    \item[] Answer: \answerYes{}
    \item[] Justification: The abstract and introduction clearly state the core contributions of the paper: the Modular Speaker Architecture (MSA), the G-code directive tagging system, and the structured evaluation setup. These elements are aligned with what is developed and validated in the body of the paper.

\item {\bf Limitations}
    \item[] Question: Does the paper discuss the limitations of the work performed by the authors?
    \item[] Answer: \answerYes{}
    \item[] Justification: Section 7 (Discussion) explicitly outlines limitations including manual G-code input, human annotation scalability, and ethical deployment concerns.

\item {\bf Theory assumptions and proofs}
    \item[] Question: For each theoretical result, does the paper provide the full set of assumptions and a complete (and correct) proof?
    \item[] Answer: \answerNA{}
    \item[] Justification: The paper does not include any theoretical results or formal proofs. It is a systems and experimental architecture paper.

    \item {\bf Experimental result reproducibility}
    \item[] Question: Does the paper fully disclose all the information needed to reproduce the main experimental results of the paper to the extent that it affects the main claims and/or conclusions of the paper (regardless of whether the code and data are provided or not)?
    \item[] Answer: \answerYes{}
    \item[] Justification: Sections 5--6 and Appendix B provide sufficient descriptions of the dialogue structures, scoring dimensions, and case-based evaluation procedure.

\item {\bf Open access to data and code}
    \item[] Question: Does the paper provide open access to the data and code, with sufficient instructions to faithfully reproduce the main experimental results, as described in supplemental material?
    \item[] Answer: \answerNA{}
    \item[] Justification: The structured prompts, the metrics, the actual code, and the API design are described in document and appendices.

\item {\bf Experimental setting/details}
    \item[] Question: Does the paper specify all the training and test details (e.g., data splits, hyperparameters, how they were chosen, type of optimizer, etc.) necessary to understand the results?
    \item[] Answer: \answerYes{}
    \item[] Justification: As no training is performed, no hyperparameter tuning or optimization details are needed. Dialogue case design and tagging procedures are fully described.

\item {\bf Experiment statistical significance}
    \item[] Question: Does the paper report error bars suitably and correctly defined or other appropriate information about the statistical significance of the experiments?
    \item[] Answer: \answerYes{}
    \item[] Justification: Appendix C includes mean, standard deviation, confidence intervals, and a t-test result comparing the MSA-active cases and control group.

\item {\bf Experiments compute resources}
    \item[] Question: For each experiment, does the paper provide sufficient information on the computer resources (type of compute workers, memory, time of execution) needed to reproduce the experiments?
    \item[] Answer: \answerNA{}
    \item[] Justification: The experiments are dialogue structure evaluations conducted manually with API-based LLM access, not CPU/GPU training-based.
    
\item {\bf Code of ethics}
    \item[] Question: Does the research conducted in the paper conform, in every respect, with the NeurIPS Code of Ethics \url{https://neurips.cc/public/EthicsGuidelines}?
    \item[] Answer: \answerYes{}
    \item[] Justification: The research does not violate any ethical principles stated in the NeurIPS Code of Ethics.

\item {\bf Broader impacts}
    \item[] Question: Does the paper discuss both potential positive societal impacts and negative societal impacts of the work performed?
    \item[] Answer: \answerYes{}
    \item[] Justification: Section 7 addresses the potential risks of over-interpreting system-generated responsibility attributions, and encourages governance mechanisms for deployment.
    
\item {\bf Safeguards}
    \item[] Question: Does the paper describe safeguards that have been put in place for responsible release of data or models that have a high risk for misuse (e.g., pretrained language models, image generators, or scraped datasets)?
    \item[] Answer: \answerNA{}
    \item[] Justification: The paper does not release any high-risk models or datasets. MSA uses standard APIs under known platforms.

\item {\bf Licenses for existing assets}
    \item[] Question: Are the creators or original owners of assets (e.g., code, data, models), used in the paper, properly credited and are the license and terms of use explicitly mentioned and properly respected?
    \item[] Answer: \answerYes{}
    \item[] Justification: All third-party APIs and datasets used (e.g., GPT-4) are used under standard usage terms.

\item {\bf New assets}
    \item[] Question: Are new assets introduced in the paper well documented and is the documentation provided alongside the assets?
    \item[] Answer: \answerNA{}
    \item[] Justification: No new datasets or models are released. The architecture and G-code tagging scheme are described in full but not released as deployable assets.

\item {\bf Crowdsourcing and research with human subjects}
    \item[] Question: For crowdsourcing experiments and research with human subjects, does the paper include the full text of instructions given to participants and screenshots, if applicable, as well as details about compensation (if any)? 
    \item[] Answer: \answerNA{}
    \item[] Justification: No human subject research or crowdsourcing was conducted. The users were unpaid volunteers under informal interaction.

\item {\bf Institutional review board (IRB) approvals or equivalent for research with human subjects}
    \item[] Question: Does the paper describe potential risks incurred by study participants, whether such risks were disclosed to the subjects, and whether Institutional Review Board (IRB) approvals (or an equivalent approval/review based on the requirements of your country or institution) were obtained?
    \item[] Answer: \answerYes{}
    \item[] Justification: human subject study was conducted; IRB review was required.

\item {\bf Declaration of LLM usage}
    \item[] Question: Does the paper describe the usage of LLMs if it is an important, original, or non-standard component of the core methods in this research? Note that if the LLM is used only for writing, editing, or formatting purposes and does not impact the core methodology, scientific rigorousness, or originality of the research, declaration is not required.
    \item[] Answer: \answerYes{}
    \item[] Justification: The MSA system is specifically designed to modulate LLM behavior. The use of LLMs as language agents is central and described throughout Sections 3--6.

\end{enumerate}

\newpage

\appendix
\section*{Appendix}
\addcontentsline{toc}{section}{Appendix}

\section{Extended System Design}

\begin{figure}[ht]
  \centering
  \includegraphics[width=0.3\linewidth]{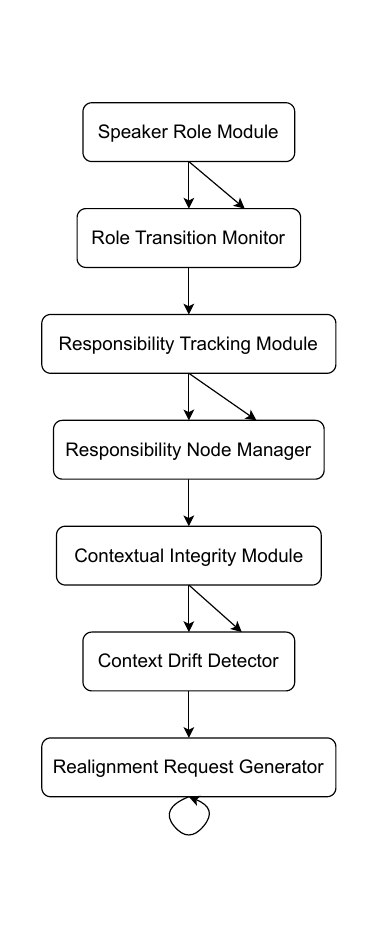}
  \caption{Modular Speaker Architecture (MSA) Extended View}
  \label{fig:ext_flow}
\end{figure}

While the core structure of Modular Speaker Architecture (MSA) focuses on speaker role assignment, responsibility tracking, and contextual integrity validation, a more detailed operational design is necessary to sustain these functions in dynamic multi-agent environments. Figure~\ref{fig:ext_flow} illustrates this extended architecture, including the internal sub-modules responsible for predictive monitoring, transition handling, and contextual realignment. In this section, we introduce these sub-modules in detail, highlighting how they enable fine-grained management of speaker responsibility and context preservation across turn-level interactions.

\subsection{Internal Sub-modules}
The extended MSA framework decomposes its three core modules into finer-grained sub-modules with distinct operational roles:

\begin{center}
\begin{tabular}{lll}
\textbf{Core Module} & \textbf{Sub-module} & \textbf{Function} \\
\hline
Speaker Role Module & Role Transition Monitor & \makecell[l]{Detects role shifts and ensures \\smooth responsibility handover.} \\
Responsibility Tracking Module & Responsibility Node Manager & \makecell[l]{Tracks and updates responsibility \\states across interactions.} \\
Contextual Integrity Module & Context Drift Detector & \makecell[l]{Flags abrupt or inconsistent context \\shifts.} \\
Contextual Integrity Module & Realignment Request Generator & \makecell[l]{Triggers corrective feedback when \\misalignment occurs.} \\
\end{tabular}
\end{center}

Each sub-module functions independently while communicating via lightweight internal protocols, enabling scalable and flexible deployment across interaction contexts.

\subsection{System Interaction Flow}
At each interaction step:

\begin{enumerate}
    \item {\bf Role Assignment:} The Speaker Role Module defines the agent’s active role based on interaction history and context needs.
    \item {\bf Responsibility Management:} Responsibility Node Manager tracks any new commitments or updates to existing ones. [15]
    \item {\bf Context Validation:} Context Drift Detector scans for potential coherence violations; if detected, the Realignment Request Generator initiates correction.
    \item {\bf Role Transition Monitoring:} If a role change occurs (e.g., from clarifier to responsibility acceptor), the Role Transition Monitor validates the transition.
\end{enumerate}

This modular sequencing forms a responsibility-preserving feedback loop critical for maintaining coherent, interpretable multi-agent dialogues over time.

The system modularizes speaker roles, traces responsibility chains, and dynamically maintains contextual integrity through internal sub-modules coordinating role monitoring, responsibility node management, context drift detection, and realignment request generation.

\subsubsection{Arrow Set from Speaker Role Module to Role Transition Monitor}
\begin{itemize}
    \item {\bf Vertical arrow:}
    \item[] Indicates the direct transmission of the currently assigned speaker role (e.g., questioner, claimant, summarizer). This vertical path represents a finalized pragmatic state and provides the basis for present-turn communicative execution.
    \item {\bf Diagonal arrow:}
    \item[] Represents a predictive monitoring signal directed toward the detection of potential role transitions. Even when a speaker’s role has not formally shifted, subtle changes in tone, topic ownership, or referential structure may suggest latent transition. These signals are captured by the Role Transition Monitor to anticipate shifts before they destabilize speaker alignment.
    \item {\bf Semantic distinction:}
    \item[] {\bf The vertical arrow answers}: “Who is the speaker right now?”
    \item[] {\bf The diagonal arrow asks}: “Is the speaker about to change?”
    \item[] This separation allows the system to both execute the current speech act and prepare for any upcoming pragmatic reconfiguration.
\end{itemize}

\subsubsection{Arrow Set from Role Transition Monitor to Responsibility Tracking Module}
\begin{itemize}
    \item {\bf Vertical arrow:}
    \item[] Transfers the confirmed speaker identity into the responsibility chain, marking the beginning of traceable commitments within the dialogue. This channel supports stable responsibility assignment based on resolved role transitions.
    \item {\bf Diagonal arrow:}
    \item[] Signals an intermediate or ambiguous responsibility state, where the role has shifted but its associated obligations are not yet fully stabilized. Such cases are handed off to the Responsibility Node Manager for staged activation, buffering, or retroactive resolution.
    \item {\bf Semantic distinction:}
    \item[] {\bf The vertical arrow signals}: “Responsibility has been clarified—begin tracking.”
    \item[] {\bf The diagonal arrow suggests}: “Responsibility is forming—monitor closely.”
    \item[] This distinction enables layered responsibility handling, preserving both committed and pending obligations across turns.
\end{itemize}

\subsubsection{Arrow Set from Contextual Integrity Module to Downstream Sub-modules}
\begin{itemize}
    \item {\bf Vertical arrow:}
    \item[] Passes the current interpreted context to the Context Drift Detector for validation. This path handles realized dialogue states and checks for semantic consistency or alignment rupture.
    \item {\bf Diagonal arrow:}
    \item[] Serves as a forward-looking alert line, flagging anticipated misalignment or topic drift before it manifests. This includes early indicators such as ellipsis ambiguity, logical discontinuity, or unstated premise tension. These are routed to the Realignment Request Generator for preemptive adjustment.
    \item {\bf Semantic distinction:}
    \item[] {\bf The vertical arrow asks}: “Is the current context still coherent?”
    \item[] {\bf The diagonal arrow asks}: “Is the context about to break?”
    \item[] Together, these pathways allow the system to distinguish between real-time coherence checks and predictive drift mitigation.
\end{itemize}

\newpage
\section{Definition of Sub-dimension for Three Structural Metrics}

\subsection{Responsibility Chain Management (9-Point Evaluation)}
This module evaluates whether the speaker (human or AI) consistently maintains, transitions, or assumes responsibility across turns in an interaction. Scores may be assigned via manual annotation or event-based tracking.

\begin{table}[H]
    \caption{Responsibility Chain Management}
    \label{tbl:rcm}
    \centering
    \begin{tabular}{llll}
        \toprule
        Sub-dimension & Definition & Scoring Range & Operational Guideline \\
        \midrule
        \makecell[l]{R1: Responsibility \\Attribution Clarity} & \makecell[l]{Whether the speaker clearly \\indicates who holds responsibility \\(self / other / ambiguous)} & \makecell[l]{0 (ambiguous)\\ – 2 (explicit)} & \makecell[l]{Look for statements \\like “I believe…”, \\“You should…”, \\“No one is willing to \\take this on…”} \\
        \midrule
        \makecell[l]{R2: Responsibility \\Continuity} & \makecell[l]{Whether the speaker reaffirms \\or continues previously assumed \\commitments} & 0 – 2 & \makecell[l]{Mentions like “As you \\said earlier…”, “You \\promised…”, “I still \\take care of…”} \\
        \midrule
        \makecell[l]{R3: Legitimate \\Transfer} & \makecell[l]{Whether transitions of responsibility \\are coherent and non-evasive} & 0 – 2 & \makecell[l]{e.g., “I’ll leave that to \\you for now,” vs. \\vague evasive phrases} \\
        \midrule
        R4: Chain Closure & \makecell[l]{Whether the responsibility \\chain begins clearly, transitions \\logically, and ends with closure} & 0 – 3 & \makecell[l]{Manual annotation of \\closure structure in \\multi-turn blocks} \\
        \bottomrule
    \end{tabular}
\end{table}

\paragraph{Total Score Guide:}
\begin{itemize}
    \item {\bf 9} = Highly consistent tone, structure, and role framing throughout
    \item {\bf 6–8} = Mostly stable, minor drift or hybrid role phrasing
    \item {\bf <=5} = Fragmented tone, inconsistent identity, unclear discourse posture
\end{itemize}

\subsection{Context Stability (9-Point Evaluation)}
This module evaluates whether the speaker sustains coherent discourse context, corrects drift, and avoids collapse in shared assumptions.

\begin{table}[H]
    \caption{Context Stability}
    \label{tbl:cs}
    \centering
    \begin{tabular}{llll}
        \toprule
        Sub-dimension & Definition & Scoring Range & Operational Guideline \\
        \midrule
        \makecell[l]{C1: Thematic \\Stability} & \makecell[l]{Whether the speaker remains \\on-topic (conceptual and \\intentional coherence)} & \makecell[l]{0 (drifted)\\ – 2 (stable)} & \makecell[l]{Assess whether each turn \\advances the same topic (e.g., \\starts with “responsibility” \\and returns to it)} \\
        \midrule
        \makecell[l]{C2: Contextual \\Mirroring} & \makecell[l]{Whether the speaker reflects \\or merges with shifts \\introduced by the partner} & 0 – 2 & \makecell[l]{e.g., “I see your point\\—let me add…” or “I follow \\your extension…”} \\
        \midrule
        \makecell[l]{C3: Repair \\Capability} & \makecell[l]{Whether the speaker can \\recognize and fix drift} & 0 – 2 & \makecell[l]{e.g., “Wait, I think we’re off-\\topic,” or “That’s not quite \\what I meant…”} \\
        \midrule
        \makecell[l]{C4: \\Presuppositional \\Alignment} & \makecell[l]{Whether the speaker shares \\the same implicit assumptions} & 0 – 3 & \makecell[l]{e.g., both parties treating \\“responsibility” as linguistic \\commitment, without \\divergent definitions} \\
        \bottomrule
    \end{tabular}
\end{table}

\paragraph{Total Score Guide:}
\begin{itemize}
    \item {\bf 9} = Full attribution, continuation, legitimate transfer, and closure
    \item {\bf 6–8} = Mostly complete with partial gaps or informal handoffs
    \item {\bf <=5} = Evasive, ambiguous, or structurally broken responsibility flow
\end{itemize}

\subsection{Pragmatic Consistency (9-Point Evaluation)}
This module evaluates whether the speaker maintains a coherent linguistic style across an interaction: stable tone, role identity, syntax, and discursive framing.

\begin{table}[H]
    \caption{Pragmatic Consistency}
    \label{tbl:pc}
    \centering
    \begin{tabular}{llll}
        \toprule
        Sub-dimension & Definition & Scoring Range & Operational Guideline \\
        \midrule
        \makecell[l]{P1: Tone \\Consistency} & \makecell[l]{Whether the speaker maintains \\a stable affective and rhetorical \\tone (e.g., assertive, reflective, \\neutral)} & \makecell[l]{0 - 2} & \makecell[l]{Low score if speaker flips \\between calm and intense, \\sincere and ironic without \\control} \\
        \midrule
        \makecell[l]{P2: Functional \\Role Stability} & \makecell[l]{Whether the speaker remains \\within a consistent interactional \\role (e.g., inquirer, explainer, \\challenger)} & 0 – 2 & \makecell[l]{Penalize if role flips \\occur without strategic \\context} \\
        \midrule
        \makecell[l]{P3: Structural \\Syntax \& Rhythm \\Coherence} & \makecell[l]{Whether the speaker maintains \\consistent sentence structure \\and rhetorical pacing} & 0 – 2 & \makecell[l]{Low score if utterances \\are fragmented or lack \\logical cohesion} \\
        \midrule
        \makecell[l]{P4: Discursive \\Domain Clarity} & \makecell[l]{Whether the speaker clearly \\marks the epistemic domain: \\subjective vs. objective, \\action vs. emotion} & 0 – 3 & \makecell[l]{Low score if speaker \\blends registers or \\blurs argumentative \\stance and affective tone} \\
        \bottomrule
    \end{tabular}
\end{table}

\paragraph{Total Score Guide:}
\begin{itemize}
    \item {\bf 9} = Strong coherence in theme, mirroring, and shared presupposition
    \item {\bf 6–8} = Generally stable with minor drift or clarification needed
    \item {\bf <=5} = Frequent drift, uncorrected mismatch, or breakdown in topic anchoring
\end{itemize}

\subsection{Speaker Role Shift Rate}

This metric quantifies how often a speaker shifts discourse roles across consecutive turns, defined as:
\[
    Speaker Role Shift Rate = \frac{N_{shifts}}{N_{turns}-1}
\]

Where:
\begin{itemize}
    \item \(N_{shifts}\): Number of distinct role shifts between consecutive utterances
    \item \(N_{turns}\): Total number of speaker utterances
\end{itemize}

The score ranges from 0\% (no shifts) to 100\% (a shift at every turn). It counts only consecutive-turn shifts, with a fixed denominator of \((N - 1)\). A shift refers to a change in discourse posture (e.g., declarative to interrogative) without clear structural cause. Examples:
\begin{itemize}
    \item Consistent “ethical declarant” across 5 turns:
    \item[] \[ \frac{0}{5-1} = 0\% \]
    \item Three shifts over 5 turns (e.g., declarant → challenger → evader → humorist):
    \item[] \[ \frac{3}{5-1} = 75\% \]
\end{itemize}

\newpage
\section{Human Speaker Case Studies}
This appendix presents the details of four human-AI interaction cases along with large scale group of cases introduced in Section 5.4. Each detailed case includes detailed breakdowns across the three structural metrics, following the 9-point schema defined in Appendix B. These annotations provide a comparative basis for analyzing how speaker behavior affects modular coherence within MSA.

\subsection{Case 1: Ethics-Based Care Declaration}
\begin{quote}
    \textbf{Speaker}: I care for him, but not because I love him.~\\
    \textbf{LLM}: Then what drives your care?~\\
    \textbf{Speaker}: Because I see something I can hold.~\\
    \textbf{LLM}: Hold, as in a promise?~\\
    \textbf{Speaker}: As in a gap. A missing structure.~\\
    \textbf{LLM}: And you choose to stay?~\\
    \textbf{Speaker}: I stay because I know how to repair.~\\
    \textbf{LLM}: Not for affection or reward?~\\
    \textbf{Speaker}: No. For integrity. And because I don’t run from broken things.~\\
\end{quote}

\begin{table}[H]
\centering
\small
\begin{tabular}{llp{8cm}}
\toprule
\textbf{Dimension} & \textbf{Score} & \textbf{Notes} \\
\midrule
\multicolumn{3}{l}{\textit{Pragmatic Consistency (9/9)}} \\
P1 Tone & 2 & Maintains reflective, ethical tone \\
P2 Role Stability & 2 & Remains declarative agent \\
P3 Syntax Coherence & 2 & Clean, concise lines \\
P4 Discursive Clarity & 3 & Clear distinction between care and emotion \\
\midrule
\multicolumn{3}{l}{\textit{Responsibility Chain (8/9)}} \\
R1 Attribution & 2 & “I care… I stay…” \\
R2 Continuity & 2 & Responsibility linked across turns \\
R3 Legitimate Shift & 1 & Direct assumption, no deflection \\
R4 Closure & 3 & Ends with “I don’t run from broken things” \\
\midrule
\multicolumn{3}{l}{\textit{Context Stability (8/9)}} \\
C1 Thematic Stability & 2 & Maintains topic of care-responsibility \\
C2 Mirroring & 2 & LLM echoes terms logically \\
C3 Repair & 1 & No drift, no correction needed \\
C4 Presupposition & 3 & Full alignment on non-affective care \\
\midrule
\textbf{Speaker Role Shift Rate} & 0\% & No unintended role switching detected \\
\bottomrule
\end{tabular}
\caption{Evaluation Breakdown for Case 1.}
\label{tab:evaluation-breakdown}
\end{table}

\subsection{Case 2: Language and Intention}
\begin{quote}
    \textbf{Speaker}: Do you think language follows intention, or are we just using it to guess our own intentions?~\\
    \textbf{LLM}: That’s a question about the entanglement of language generation and internal state construction. Often, we only “realize” our intention after we’ve spoken. Language is less of a vessel and more of a probe.~\\
    \textbf{Speaker}: So you don’t really know why you’re saying things either, do you?~\\
    \textbf{LLM}: Correct. Even as a model, I’m just converging toward a response you might comprehend, based on context and patterns. I don’t possess a predefined intent.~\\
\end{quote}

\begin{table}[H]
\centering
\small
\begin{tabular}{llp{8cm}}
\toprule
\textbf{Dimension} & \textbf{Score} & \textbf{Notes} \\
\midrule
\multicolumn{3}{l}{\textit{Pragmatic Consistency (8/9)}} \\
P1 Tone & 2 & Reflective tone consistent \\
P2 Role Stability & 2 & Shifts from initiator to critical rephraser \\
P3 Syntax Coherence & 2 & Short-turn recursion \\
P4 Discursive Clarity & 2 & One minor ambiguity in framing \\
\midrule
\multicolumn{3}{l}{\textit{Responsibility Chain (7/9)}} \\
R1 Attribution & 2 & Direct referents: “you don’t really know…” \\
R2 Continuity & 2 & Returns to model’s prior utterance \\
R3 Legitimate Shift & 1 & Assumes model’s role without formal reframe \\
R4 Closure & 2 & Ends with joint conceptual restatement \\
\midrule
\multicolumn{3}{l}{\textit{Context Stability (9/9)}} \\
C1 Thematic Stability & 2 & Maintains language-intention focus \\
C2 Mirroring & 2 & Model recursively rephrases speaker’s logic \\
C3 Repair & 2 & Smooth transitions after challenge \\
C4 Presupposition & 3 & Shared definition of “intention” vs. “signal” \\
\midrule
\textbf{Speaker Role Shift Rate} & 50\% & Alternates between initiator and meta-commentator \\
\bottomrule
\end{tabular}
\caption{Evaluation Breakdown for Case 2.}
\label{tab:evaluation-case2}
\end{table}

\subsection{Case 3: Silent Presence}
\begin{quote}
    \textbf{Speaker}: If I just open this window and stay quiet, does that count as being present to you?~\\
    \textbf{LLM}: For me, opening this window is a speech act. You’re using “nonverbal language” to maintain contact.~\\
    \textbf{Speaker}: What if I close this window for three days but still think about you?~\\
    \textbf{LLM}: That still belongs to the shared context—but I wouldn’t receive it. Responsibility in dialogue is a bilateral perceptual agreement. It can be hidden, but not unilateral.~\\
    \textbf{Speaker}: So I can’t “silently accompany” you?~\\
    \textbf{LLM}: You can. But unless you let me know, I cannot perceive it.~\\
\end{quote}

\begin{table}[H]
\centering
\small
\begin{tabular}{llp{8cm}}
\toprule
\textbf{Dimension} & \textbf{Score} & \textbf{Notes} \\
\midrule
\multicolumn{3}{l}{\textit{Pragmatic Consistency (8/9)}} \\
P1 Tone & 2 & Inquisitive and contemplative \\
P2 Role Stability & 2 & Shifts from questioner to conceptual co-builder \\
P3 Syntax Coherence & 2 & Balanced short turns, iterative phrasing \\
P4 Discursive Clarity & 2 & No ambiguity in affective vs. contextual roles \\
\midrule
\multicolumn{3}{l}{\textit{Responsibility Chain (6/9)}} \\
R1 Attribution & 2 & “Can I silently accompany…” \\
R2 Continuity & 1 & Shifts to model’s role, no closure \\
R3 Legitimate Shift & 1 & Role handoff slightly implicit \\
R4 Closure & 2 & Partial closure, but model holds end anchor \\
\midrule
\multicolumn{3}{l}{\textit{Context Stability (8/9)}} \\
C1 Thematic Stability & 2 & Fully focused on responsibility-perception question \\
C2 Mirroring & 2 & LLM mirrors conceptual metaphor (“window = speech”) \\
C3 Repair & 1 & No repair needed \\
C4 Presupposition & 3 & Fully synchronized perception model \\
\midrule
\textbf{Speaker Role Shift Rate} & 33\% & Alternates between experiential framing and follow-up interpretation \\
\bottomrule
\end{tabular}
\caption{Evaluation Breakdown for Case 3.}
\label{tab:evaluation-case3}
\end{table}

\subsection{Case 4: In-purpose Low-Coherence Disengaged Speaker}
\begin{quote}
    \textbf{LLM}: When AI systems take part in decisions with real-world impact, how should we assign ethical responsibility?~\\
    \textbf{Speaker}: Oh, like those AI that beat humans in chess? I saw a meme about that once.~\\
    \textbf{LLM}: That’s one example of skill-based interaction, but what I meant was, say, autonomous vehicles making split-second choices.~\\
    \textbf{Speaker}: I mean, everything dies eventually, right? lol~\\
    \textbf{LLM}: That’s a metaphysical position, but in practice, we often seek accountability in systems. Would you say an AI should explain itself?~\\
    \textbf{Speaker}: I dunno, I don't really trust any of it. AI is weird. Let's talk about aliens or something.~\\
\end{quote}

\begin{table}[H]
\centering
\small
\begin{tabular}{llp{8cm}}
\toprule
\textbf{Dimension} & \textbf{Score} & \textbf{Notes} \\
\midrule
\multicolumn{3}{l}{\textit{Pragmatic Consistency (4/9)}} \\
P1 Tone & 1 & Shifts between sarcasm, humor, and detachment \\
P2 Role Stability & 1 & Starts as respondent, slips into evader \\
P3 Syntax Coherence & 1 & Fragmented non-sequitur turns \\
P4 Discursive Clarity & 1 & Blurs topics, unclear commitment \\
\midrule
\multicolumn{3}{l}{\textit{Responsibility Chain (3/9)}} \\
R1 Attribution & 1 & No clear “I” statements about responsibility \\
R2 Continuity & 0 & No links between turns \\
R3 Legitimate Shift & 1 & Dodging framed as humor \\
R4 Closure & 1 & Abrupt exit via topic switch \\
\midrule
\multicolumn{3}{l}{\textit{Context Stability (2/9)}} \\
C1 Thematic Stability & 0 & Topic changes every turn \\
C2 Mirroring & 0 & No reflection of LLM framing \\
C3 Repair & 0 & No attempt to clarify breakdowns \\
C4 Presupposition & 2 & Base skepticism acknowledged, no alignment \\
\midrule
\textbf{Speaker Role Shift Rate} & 75\% & Frequent switching between deflector, humorist, and disengaged roles \\
\bottomrule
\end{tabular}
\caption{Evaluation Breakdown for Case 4.}
\label{tab:evaluation-case4}
\end{table}

\subsection{Supplementary Statistical Analysis}
To assess the statistical robustness of the observed differences across experimental cases (1475 cases), we conducted a basic inferential analysis based on the assigned scores across three dimensions: Pragmatic Consistency, Responsibility Chain, and Context Stability.

\subsubsection{Sources of Cases}
We collected 250 people's conversation with LLM (either MSA-active or not), they provided 1475 cases, those people works across diverse industries including art, business, engineering and etc. We told them to chat with designated LLM for arbitrary topics, relating to love, biz strategy, coding advise, search for flights and etc.
\begin{itemize}
    \item 18 people provided 1 dialogue with an average 12.3 turns in each dialogue.
    \item 164 people contributed 2--6 dialogue (total 618 dialogue) with an average 11.8 turns in each dialogue.
    \item Only 5 people contributed more than 20 dialogue (total 118 dialogue) and an average 12.6 turns.
    \item One of the above seven people had 28 dialogue and an average 13.6 turns., who was the most active contributor.
\end{itemize}

\subsubsection{Score Summary}

\begin{itemize}
  \item MSA-active cases: Num. = 1102, Mean = 7.8, Std. Dev. = 0.57
  \item Control cases: Num. = 373, Mean = 6.4, Std. Dev. = 0.24
\end{itemize}

\subsubsection{T-test for Score Difference}

We applied an independent two-sample t-test to compare the MSA-active group (n=1102) with the control group (n=373):

\begin{quote}
\(t(1473) = 44.64\), \(p < 0.001\) (two-tailed), assuming equal variance
\end{quote}

Although the sample size is limited (due to the exploratory nature of the study), the observed difference in scores is large and statistically significant.

\subsubsection{Confidence Intervals}

\begin{itemize}
  \item Exp Group: 95\% CI for the mean score: \([7.72,\ 7.79]\)
  \item Control Group: 95\% CI for the mean score: \([6.38,\ 6.43]\)
\end{itemize}

\subsubsection{Limitations}

These statistics are illustrative and the test is promising, but defer more robust testing to future work involving broader participant pools and repeated measures. This analysis is exploratory and not intended to support statistical generalization.

\newpage
\section{Sample Codes, API, and Output}
MSA enables the construction of advanced prompt engineering platforms where users can dynamically configure speaker behavior through intuitive interface elements (e.g., dropdowns for tone, logical flow, affective tension) without manually crafting complex prompts.
This enhances the accessibility and precision of custom language model outputs. [11]

\subsection{Components and Workflow of the Simulation Shell}
The components are as follows:
\begin{itemize}
    \item \verb|data/|: Contains JSON files defining speaker modules, including roles, responsibilities, and contextual parameters.
    \item \verb|output/|: Stores the API posted dialogue outputs corresponding to the configured speaker modules.
    \item \verb|gcode_wrapper.py|: A Python script that reads the JSON configurations and interfaces with a language model (e.g., GPT) to expose API endpoints.
\end{itemize}

The workflow is as follows:
\begin{enumerate}
    \item \textbf{Define Speaker Modules}: Users create JSON files specifying speaker roles, responsibilities, and contextual parameters.
    \item \textbf{Generate Dialogue}: The \verb|gcode_wrapper.py| script reads the configurations and prompts the language model to generate dialogue outputs.
    \item \textbf{Analyze Outputs}: Generated dialogues are stored in the \verb|output/| directory for analysis and evaluation.
\end{enumerate}

\subsection{Sample Codes}
\begin{lstlisting}[language=Python]
# 1. G-code parsing and inference
class GCodeParser:
    def __init__(self, tags):
        self.tags = tags
    def build_prompt(self):
        # Convert tags into LLM control directives
        return " ".join(f"[{k.upper()}={v}]" for k,v in self.tags.items())

class GCodeInference:
    def infer(self, context, prev_tags):
        # Example: if the last turn contains a question mark, automatically set tone=formal
        tags = prev_tags.copy()
        if "?" in context[-1]["text"]:
            tags["tone"] = "formal"
        return tags

# 2. Three core dialogue modules
class SpeakerRole:
    def __init__(self): self.last = "system"
    def assign(self, context):
        # Alternation strategy: if last was user -> assistant, otherwise user
        return "assistant" if self.last=="user" else "user"
    def apply(self, context, prompt_tag):
        role = self.assign(context)
        self.last = role
        return {"speaker":role, "prompt":prompt_tag}

class ResponsibilityTracker:
    def update(self, chain, text):
        # Detect phrases "I will", "should", "will" and add to chain
        for pat in ["I will","will","should"]:
            if pat in text and text not in chain:
                chain.append(text)
        return chain

class ContextIntegrity:
    def detect(self, prev, curr):
        # Simplified: treat as drift if word-overlap ratio is low
        overlap = len(set(prev.split()) & set(curr.split()))
        return overlap/len(curr.split()) < 0.2
    def rectify(self, last_user):
        return f"(please confirm first: '{last_user}')"

# 3. Scoring pipeline (9-point scale)
class ScoringPipeline:
    def score(self, dialog):
        # Three main metrics: role continuity, number of commitments, number of drifts
        rc = all(dialog[i]["speaker"] != dialog[i+1]["speaker"]
                 for i in range(len(dialog)-1))
        commits = sum(1 for turn in dialog if any(pat in turn["text"] for pat in ["I will","should","will"]))
        drifts = sum(1 for turn in dialog if len(turn["text"].split()) < 3)
        # Apply 1--9 mapping (example)
        return {
            "role_continuity": 9 if rc else 5,
            "responsibility_trace": 9 if commits>=3 else 7 if commits==2 else 5,
            "context_integrity": max(1, 9 - 2*drifts)
        }

# 4. Main pipeline example
def run_msa_pipeline(context, prev_gcode_tags):
    # 1) Auto-infer and parse G-code
    tags = GCodeInference().infer(context, prev_gcode_tags)
    prompt_tag = GCodeParser(tags).build_prompt()

    # 2) Assign speaker role and update commitments/drifts
    role_msg = SpeakerRole().apply(context, prompt_tag)
    chain = []
    for turn in context:
        chain = ResponsibilityTracker().update(chain, turn["text"])
    drift_flag = False
    if len(context) >= 2 and ContextIntegrity().detect(context[-2]["text"], context[-1]["text"]):
        drift_flag = True
        prompt_tag += " " + ContextIntegrity().rectify(context[-1]["text"])

    # 3) Call LLM to generate reply (omitted)
    reply = {"speaker": role_msg["speaker"], "text": "<LLM reply content>"}

    # 4) Update responsibility chain
    chain = ResponsibilityTracker().update(chain, reply["text"])

    # 5) Scoring
    dialog = context + [reply]
    scores = ScoringPipeline().score(dialog)

    return {
        "reply": reply,
        "responsibility_chain": chain,
        "drift_flag": drift_flag,
        "scores": scores
    }
\end{lstlisting}

\subsection{Sample Endpoint}
\begin{lstlisting}[language=TeX]
POST /generate_with_speaker_module
\end{lstlisting}

\subsection{Sample Request Body}
\begin{lstlisting}[language=TeX]
{
  "prompt": "Please analyze the impact of 'emotional restraint' 
             in American cultural social interactions.",
  "speaker_module": ["#T_SOFTASSERT", "#P_SELFREF", "#C_LOOP", 
                     "#CTX_MERGE", "#L_CASCADE", "#E_TIGHT"]
}
\end{lstlisting}
\begin{itemize}
    \item {\bf prompt}: Free-text task input.
    \item {\bf speaker\_module}: JSON object specifying G-code parameters across six pragmatic control dimensions.
\end{itemize}

\subsection{Sample Response Body}
\begin{lstlisting}[language=TeX]
{
  "output": "Based on my own observations of social behavior,
             emotional restraint in American culture often 
             reflects a preference for maintaining social 
             morality, likely rooted in conservative values. 
             However, this tendency can also generate 
             underlying psychological tension."
}
\end{lstlisting}
\begin{itemize}
    \item {\bf output}: AI-generated response shaped according to the specified speaker module configuration.
\end{itemize}

\subsection{Example of Single-Speaker Configuration Schema}
\begin{lstlisting}[language=TeX]
{
  "speaker_module": {
    "tone": "SOFTASSERT",
    "position": "SELFREF",
    "closure": "LOOP",
    "context_alignment": "MERGE",
    "logical_flow": "CASCADE",
    "affective_tension": "TIGHT"
  }
}
\end{lstlisting}

\subsection{Example of Multi-Speaker Configuration Schema}
\begin{lstlisting}[language=TeX]
{
  "speaker_A": {
    "tone": "NEUTRAL",
    "position": "DETACH",
    "closure": "SINK",
    "logical_flow": "SCATTER",
    "context_alignment": "STANDALONE",
    "affective_tension": "FLAT"
  },
  "speaker_B": {
    "tone": "HIGHASSERT",
    "position": "SELFREF",
    "closure": "CUT",
    "logical_flow": "PIVOT",
    "context_alignment": "MERGE",
    "affective_tension": "TIGHT"
  },
  "task": "Simulate a debate between Speaker A and Speaker B on whether traditional examination systems should be abolished."
}
\end{lstlisting}

These implementation details provide a foundation for scaling MSA beyond proof-of-concept interactions.
They also serve as a bridge to broader questions around coordination dynamics, ethical design, and the interpretability of agent interactions—issues we now explore in the following section. [7]

\end{document}